\def\year{2021}\relax
\documentclass[letterpaper]{article} 
\usepackage{aaai21}  
\usepackage{times}  
\usepackage{helvet} 
\usepackage{courier}  
\usepackage{wrapfig}

\usepackage[hyphens]{url}  
\usepackage{graphicx} 
\usepackage{natbib}  
\usepackage{caption} 
\title{Explanation from Specification}
\author {
    Harish Naik,\textsuperscript{\rm 1}
    Gy\"{o}rgy Tur\'{a}n\footnote{Partially supported by the National Research, Development and Innovation Office of Hungary through the Artificial Intelligence National Excellence Program (grant no.: 2018-1.2.1-NKP-2018-00008).} \textsuperscript{\rm 1,\rm 2} \\
}
\affiliations {
    \textsuperscript{\rm 1} University of Illinois at Chicago \\
    \textsuperscript{\rm 2} MTA-SZTE RGAI, Szeged\\
    \{hnaik2,gyt\}@uic.edu, 
}

\begin{document}

\maketitle

\begin{abstract}
 Explainable components in XAI algorithms often come from a familiar set of models, such as linear models or decision trees.
We formulate an approach where the type of explanation produced
 is guided by a specification.
Specifications are elicited from the user, possibly using interaction with the user
and contributions from other areas.
Areas where a specification could be obtained include forensic, medical and scientific applications. Providing a menu of possible types of specifications in an area is an exploratory knowledge representation and reasoning task for the algorithm designer, aiming at understanding the possibilities and limitations of efficiently computable modes of explanations.
Two examples are discussed: explanations for Bayesian networks using the theory of argumentation, and explanations for graph neural networks. The latter case illustrates the possibility of 
having a representation formalism available to the user for specifying the type of explanation requested, for example, a chemical query language for classifying molecules. 
The approach is motivated by a theory of explanation in the philosophy of science,
and it is related to current questions in the philosophy of science on the role of machine learning.
\end{abstract}

\section{Introduction}

Interpretability, explainability, transparency~\footnote{In the introductory discussion we use ``interpretability'' for the other related notions as well. The rest of the paper deals with explainability.}
and other requirements for neural networks and other models built using machine learning pose an important challenge in many applications.
The history of the problem in machine learning is illustrated by \cite{Sommer96}: ``This paper reopens the issue of understandability of induced theories, which, while prominent in the early days of ML, seems to have fallen out of favor in the sequel.''
It is often noted that these notions are used with different meanings and precise definitions do not exist (see, e.g., \cite{Lipton18}). Possibly the notions are inherently ambiguous, and therefore giving formal definitions is not necessary and perhaps not even possible.
Precise definitions may not be of interest in themselves, but they can be useful, even in restricted cases, for understanding the possibilities and limitations of algorithmic approaches.

In view of the large variety of algorithms for interpretability, several papers provide {\em taxonomies} (e.g., \cite{Arrieta20,Guido19,HeninMe,Sokol20}). These are not formal definitions, but describe various components of the algorithms, different types of those components and their interactions, objectives and  evaluation methods.  
\cite{DoshiUjj} give a taxonomy of approaches for evaluating interpretability.
The main criterion for evaluating interpretability of a result is its usefulness for humans.

\cite{MillerInm}, quoting \cite{CooperBook}, 
 refer to AI researchers designing interpretable learning algorithms without taking into account desiderata based on philosophy, cognitive science and other areas as ``inmates running the asylum''.
While this suggests that computer scientists have too large a role, it may also be the case that their task is narrow in the sense that interpretable models usually are of a familiar kind, such as linear models and decision trees. Are there other options? We formulate an approach to post-hoc explainability which, while adhering to Miller's warning, could help answer this question.

The proposed approach is that for an XAI application, instead of deciding on the explainable model to be used (say, a decision tree), one can try to obtain a \emph{specification} of the kind of explanation required.
The task then is to design an explainable learning model satisfying the requirements, or to argue that this is not possible, for informational or computational reasons.
This view is related to the taxonomy of \cite{DoshiUjj}. 
Another view is that the specification represents the type of {\em question} or {\em query} and the type of {\em answer} expected by the user, providing an abstraction of the user
for the  application.

It is to be emphasized that 
the approach is not expected to be feasible in every application, but it is hoped that in some cases it could
lead to the development of new explainable representations.
It could also provide an understanding of the relevant features of explainable models or, in the specification phase, provide a better understanding of the explainability requirements of the particular application. Such an approach is feasible in \emph{forensic, medical} and \emph{scientific} applications, where the user has an approximate idea of what is meant by an explanation. This approximate idea can then be refined in an interaction with the algorithm designer.

The algorithm designer can prepare for such an interaction by studying various types of explanations that are to be expected in a particular area.
Thus the proposed framework includes exploring knowledge representation and reasoning aspects of explainability.
For a given type of explanation, an alternative route is to learn a corresponding interpretable representation directly. The feasibility of this route can depend on the nature of data available for a particular application.

In this paper we give several motivations for explanations from specifications, including functionally grounded evaluation of explainability methods, the pragmatic theory of explanations and the theoretical study of explainability. We describe several aspects of the realization of the approach, and we present two examples
where there is initial work in the direction proposed: explaining Bayesian network classifiers using argumentation theory and explaining graph neural network. Future work is outlined in both cases. The philosophy of science gives a motivation for the approach, and we also mention a connection in the other direction as well, noting that the exploration of the approach could contribute to the discussion of current problems on the role of machine learning in scientific research.


\section{Functionally grounded evaluation} \label{sec:fun}
In the taxonomy of \cite{DoshiUjj} for interpretability evaluation methods, functionally grounded evaluation does not use human
experiments. Instead, a
``formal definition of interpretability [is used] as a proxy for explanation quality [...] [which is] most appropriate once we have a class of models or regularizers that have already been validated [...] The challenge, of course, is to determine what proxies to use. For example, decision trees have been considered interpretable in many situations''.

A project in this direction, proposed in \cite{DoshiUjj}, is to create a matrix with rows corresponding to real-world tasks, columns corresponding to methods, and entries corresponding to the performance of methods on the task (like decision trees of a certain type for a medical interpretability task). An interesting hypothesis formulated in \cite{DoshiUjj} is that such a matrix may be factored using latent interpretability requirements of tasks such as global versus local interpretability and the type of user expertise required, and latent properties of the methods, e.g., the structure of methods in terms of cognitive chunks.
The approach proposed in this paper {\em promotes} these factors to a central role and makes their 
elicitation a separate task guiding algorithm design.
\section{The pragmatic theory of explanation} \label{sec:pra}
Interpretability and explainability are much studied, distinct but related, notions in philosophy.
Scientific explanation, in particular, is a central concept in the philosophy of science.
Several approaches are formulated, starting with the deductive-nomological (DN) approach of \cite{Hempel}. The approaches are quite different and there is no generally accepted one.
Whether this background is relevant for ML is debated~\cite{Krishnan19,Paez19}~\footnote{The relationship between scientific explanation and explanation in general is also open to debate.}.
One possible answer is
that approaches in the philosophy of science are of interest in AI as potentially useful views of what constitutes an explanation in ML. This position involves no commitment on the relationship between the corresponding notions in the philosophy of science and AI. 

One difference to note is that the objective in the philosophy of science is to capture the notion of a scientific explanation precisely. 
Counterarguments against a particular approach may refer to capturing too much or too little. 
In AI, on one hand, the bar is lower, as useful approaches are sought for, without any claim of precise characterization.
On the other hand, the bar in AI is higher, because the approaches are required to be algorithmically tractable.
 The distinction between explaining the learned model or explaining the phenomenon modeled is an important one.
 Robustness may provide a distinction between the two~\cite{Jaakkola5,Hancox}.


The approach proposed in this paper is inspired by the {\em pragmatic theory of explanation} of \cite{FraassenBook}~\footnote{Here ``pragmatic'' does not refer to claim of relevance for practice (which would apply to other approaches as well), but to aspects of the participation of the agents in the explanation process (the details of which are not needed for our discussion).
See the distinction between pragmatic$_1$ and pragmatic$_2$ in~\cite{Woodward}.}. This theory is part of van Fraassen's constructive empiricist approach to the philosophy of science.
The main aspects are that 
explanations are viewed as {\em answers to why-questions} in a {\em contrastive form} in the context of a {\em relevance} relation
~\footnote{\cite{LiptonP} cites van Fraassen as one of the sources of the contrastive approach.
 The critical review of \cite{Worrall} on van Fraassen's book argues, among other things, that the contrastive approach is not applicable to scientific explanation, and context and relevance relations ``lead to unnecessary complexity'' by moving beyond classical logic. Lipton gives a counter-argument to Worrall's first argument. Regarding the second argument, Worrall is actually right about complexity (whether the move is necessary or not is another question). Non-classical logics, like non-monotonic logic and argumentation theory, do have problems with intractability, and overcoming those is an important problem. See further discussion in the section on argumentation.}. 
 As summarized in \cite{CrossR},
``to ask why is to ask for a reason, and [the relevance relation] $R$ varies according to the kind of reason that is being requested in a given context. One can ask why {\em in order to request causal factors, to request a justification, to request a purpose, to request a motive, to request a function, and so on''}.

One objection to van Fraassen's approach is that it is not specified what counts as a relevance relation.
We describe briefly the definition in the version given by \cite{Boniolo}. 
A {\em why-question} is of the form ``why $P$ rather than the other elements of $X$'', where $P$ is a proposition and $X$ is a set of propositions containing $P$.
The question also specifies
\begin{description}
\item a knowledge base $K$, a scientist $s$ and the {\em background knowledge} $K_s \subseteq K$ of $s$,
\item a {\em context} $C_s \subseteq K_s$ containing a {\em relevance relation} $R$, and possibly other items,
 \end{description}
where $R$ is binary relation between an {\em answer} $A$ and pairs of the form $\langle P, X \rangle$.
For a proposition $A$, ``{\em Because $A$}'' is an {\em answer} to the why-question if $R(A, \langle P, X \rangle)$ holds,
$A$ is valid and plausible w.r.t. $C_s$, and $P$ is more valid and plausible w.r.t. $C_s$ than the other members of $X$.



In AI terms there is a domain with a given background knowledge and an agent with its own background knowledge. The context the agent considers
contains its input and the kind of ``reasoning'' the agent is using. Furthermore, the input has an output $P$, and the question is: why is $P$ the output in contrast to the other possibilities in $X$? Here ``reasoning'' is used as a suggestive term for the relevance relation, which is not restricted to any kind of logical or causal reasoning and no notion of truth is involved.
\section{The proposed approach} \label{sec:app}
\begin{figure}
    \centering
    \includegraphics[scale=0.8,keepaspectratio=true]{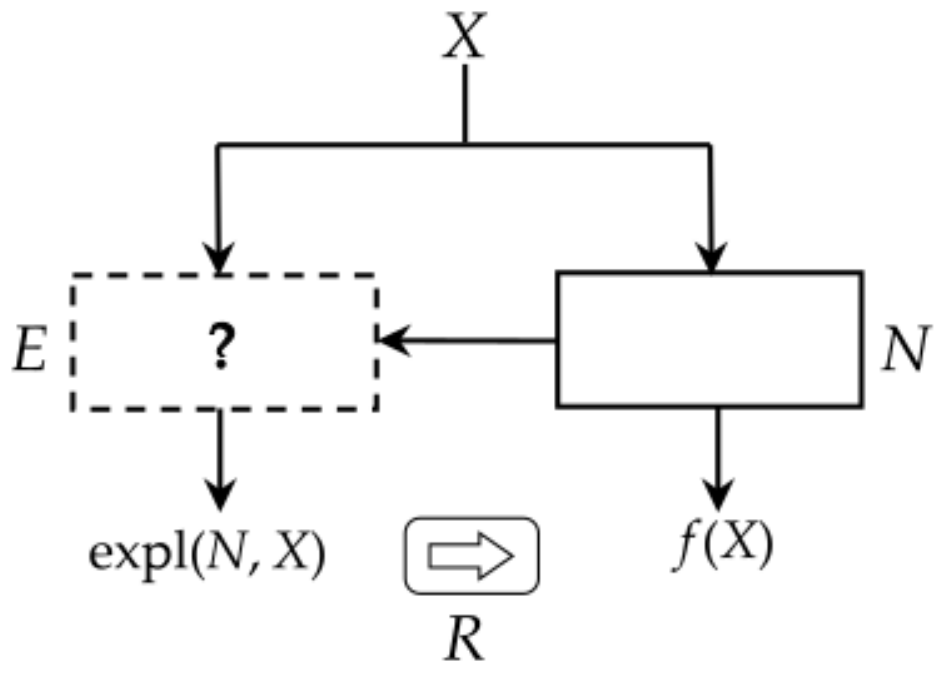}
    \caption{Diagram of the proposed approach}
    \label{fig:exp_model}
\end{figure}

The proposed view of explainability is summarized in  Figure~\ref{fig:exp_model}. 
The $N$ box represents a black box model ($N$ for network, neural or Bayesian), which on a given input $X$ produces output $f(X)$.
The $R$ module ($R$ for relevance or reasoning) represents the user's notion of an explanation. The missing component with the question mark is $E$ (the explainer), producing an explanation $expl(N, X)$ which explains $f(X)$ according to the user's requirements. The task of the algorithm designer is to design an appropriate explainer. It is assumed that the designer knows $R$, and has to develop an explainer module which, given an $N$ and an $X$, produces a suitable explanation for $f(X)$.



The contrastive aspect is reduced here to ``why $f(X)$ and not something else'' for simplicity, but the general version could also be incorporated.
The approach includes other scenarios as well, such as building an interpretable model directly or computing explanations from $N$ directly.
The setup allows for the design of $N$ to take into consideration $R$,
for the incorporation of explanation construction into the learning process~\cite{Park18},
for methods where the learning process of $N$ and the production of explanations are intertwined~\cite{Shedivat}, and, depending on $R$, for interaction as well.

The diagram, mutatis mutandis, is also related to other tasks such as the {\em verification} \cite{Katz17} or {\em testing}~\cite{Harman} of black boxes.
\cite{Dhurandhar35} gave a formalization of the user in terms of a {\em target model}, the performance of which is to be improved using interpretability.
\cite{OvertonE,Overton11} discusses the formalization of various kinds of explanations in science.

Being built around the relevance relation $R$, the approach implements van Fraassen's theory of explanation.
 By moving part of the problem formulation to the user and leaving ``only'' algorithm design
to the computer scientist, the {\em ``can is kicked down the road"}.
Furthermore, 
by building on the relevance relation and the context of the user, the approach is ``as domain-specific as it gets'', thereby adhering to the {\em domain-specificity} of interpretability~\cite{RudinNMI}.

\cite{Guido19} consider an explanation to be an interface between the decision maker (the black box) and the user, which is an accurate proxy of the decision maker and is comprehensible to the user.
The diagram suggests a way
to achieve the simultaneous occurrence of the two conditions (accurate proxy and comprehensible) required for explanations to be useful.
The user is supposed to provide $R$, which, by definition, is comprehensible to them. This could include statistical or logical aspects, background knowledge, etc.
Then the designer has to come up with the explainer, which is an accurate {\em local} proxy in the form of the requested type of explanation, which can refer to either the model or the world.

The relevance relation should be formulated in cooperation with the user, building on psychology, cognitive science,
HCI and other areas. 
This challenge seems to be somewhat different from tasks such as eliciting user preferences in terms of parameters of decision trees.
It is interesting that the importance of {\em contrastive explanations}, emphasized by \cite{Miller19} is already `built in'' into van Fraassen's framework.
It is noted by \cite{Miller19} that eliciting a contrast case from a human observer may be a difficult problem.
The present elicitation task may not be easy either, but it may 
be useful in other contexts as well.
Elicitation and related aspects are described in \cite{Cassens:2007,LimDA09,Kaur20}.

The approach can be relevant for forensic, medical and scientific applications, as there are relevance relations
in these areas which can serve as specifications.
Although it is not indicated in the figure above, 
the use of \emph{background knowledge} is especially important in these areas. Background knowledge can be incorporated explicitly and implicitly in different ways, including the design of neural networks (e.g., architecture, initialization and regularization) and the construction of explanations.
Recent surveys of work relevant for these aspects are \cite{Ruden19} on uses of background knowledge, \cite{KarpatneA} on theory-guided data science and \cite{Roscher20} on explainability aspects of current science research.

In biology there is a large variety of explanations, including non-causal forms of explanation as well, such as functional and evolutional explanation (see~\cite{Braillard}). Boniolo, in his paper cited above, gave a taxonomy of relevance relations for explanations in biology. Relevant types of explanation, besides those mentioned by \cite{Boniolo}, are also given in \cite{Trujillo}. Their MACH (methods, analogies, context, how) model is obtained from interviews and data from experts in the context of biology education, and thus it gives an example of the elicitation process. \cite{Overton12,Overton13} gives a theory of explanation in science, and relates it to an analysis of papers from {\em Science}. He uses the categories {\em theory, model, kind, entity} and {\em data}. For every pair there is a {\em core} relation for explanations involving that pair. These relations are further candidates to explore in the role of the $R$ box.

\section{Explanations for Bayesian networks using argumentation} \label{sec:arg}
Explanations for Bayesian networks have been studied for a long time~\cite{LacaveD02}.
The issue of interpretability of Bayesian networks requires some clarification. 
Bayesian networks are a clear and interpretable 
representation of dependencies in the joint distribution 
(although even that requires care to formulate exactly in terms of distribution-independence). 
On the other hand, inference given some evidence is computationally
involved and hard to interpret
for users not familiar with details of probabilistic reasoning.

Koller~\cite{Ford18} noted that such models have an intermediate degree of interpretability.
This intermediate position suggests that interpretability of probabilistic models may be more accessible than deep learning and may be of interest for exploring general issues.

{\em Argumentation} is a basic form of human reasoning~\cite{Mercier11}. important in legal and medical contexts.
{\em Argumentation theory} is a logic formalism for 
the mechanism of arriving at a 
decision making based on arguments for or against a decision, using relations like defeat, rebuttal and undercut between arguments~\cite{PollockBook}. The framework of \cite{Dung95} is a directed graph with the arguments as vertices and directed edges corresponding to the {\em attack} relation. This abstract framework is applied to a class of formulas by defining the attack relation over arguments built from those formulas. 

Argumentation has been used for explanations both directly and with neural networks (see, e.g., the survey~\cite{KakasAA}, and \cite{Coca8}). These applications build on tractable versions of the general framework.

Motivated by forensic applications, 
\cite{Timmer17} considered argument-based explanations for Bayesian network inference.
In terms of the approach proposed above, in this case a why-question is a {\em request for arguments}.
Here the $N$ box is a Bayesian network, $X$ is the evidence, the relevance relation $R$ is
implication in some argumentation framework, and the explainer $E$
is an algorithm producing (probabilistic) 
arguments replacing network inference.
The algorithm produces efficiently computable arguments for a class of networks, giving faithful explanations. 
\cite{Prakken17} analyzed the translation further, identified desirable properties, and also several undesirable ones in terms of the general theory of probabilistic argumentation frameworks.

The elicitation of $R$ involves the experimental study of argumentation frameworks. This is a recent direction of research 
\cite{Cerutti14,Cramer19}. 
The {\em grounded extension} semantics in \cite{Timmer17} is not a particularly good match. 
A relevant conclusion of 
\cite{Cerutti14} is that domain specific knowledge needs to be considered.
Taking more forms of background knowledge into account is also a consideration for further work on \cite{Timmer17}.

Thus further work on argument-based Bayesian network inference explanation needs the cooperation of algorithm designers,  cognitive scientists and users.  It requires further work to identify desirable characteristics of the argument frameworks, followed by the design, if possible, of efficient algorithms producing arguments with those characteristics, making use of background knowledge. 

\section{Explanations for graph neural networks} \label{sec:gra}

The graph neural network (GNN) model is a neural network variant for problems on graphs \cite{Hamilton5}. Nodes in each layer correspond to vertices of the input graph. Computation and updating between layers are done along the edges of the graph, giving a deep learning version of message passing algorithms. 
GNN can be used, for example, for classifying nodes of a graph, or for classifying graphs.

GNN are an important tool for classifying molecules for problems in chemistry, biology and drug design. In these areas it is often noted that even though significant progress has been made by deep learning
in terms of prediction, lack of interpretability of the results is a major problem to solve (see, e.g.,~\cite{Ching14}).
Providing explanations for node and graph classification has been studied recently by \cite{Preuer19,GNNEx,Pope19,Azizpour,Huang45}.


Properties of molecules are often studied by looking for relevant substructures. The GNNExplainer method of~\cite{GNNEx} finds explanations for the classification of a graph by a GNN in the form of a size-bounded subgraph and a subset of its node features that have large mutual information with the output. 
Thus in this case $N$ is a neural network, $X$ is a graph, an explanation is a subgraph, and the relevance relation is the subgraph relation.

\cite{Jimenez} mention the development of new interpretable molecular representations as another major research problem. 
There are many different representations and explanation types for molecules, even for a single property (see, for example, the literature on aromaticity~\cite{Stanger,Sola57}).
\cite{Barzilay5} investigate the predictive power of deep learning based representations when compared to fixed molecular input descriptions. A comparison with respect to explainability would also be interesting.
In chemoinformatics there are several query languages describing properties of molecules, such as SMILES, SMARTS and CSRML~\cite{Tarkhov}.

GNN learning algorithms and explainability methods are often evaluated on \emph{synthetic} graph data as well. Synthetic problems appear to be more realistic approximations of real-life properties for molecules than, for example, for image recognition. Therefore it seems to be of interest to explore synthetic problems in this context in more detail.

Our approach suggests that scientists can use classes of queries in the query languages to specify the type of explanation that is meaningful to them.
As an exploratory research project, one can consider
various classes of queries as candidate explanation types for synthetic problems, and study the possibilities and limitations of producing explanations of those types. 
Going beyond substructures, one could consider more general queries.
Extensions could include, for example, properties involving the existence of multiple copies of substructures (perhaps in specified positions like being far away from each other), Boolean combinations or properties involving quantification and counting.

As noted in the introduction, for each such query class one could consider learning such a description directly. 
GNN are known to have theoretical computational limitations~\cite{Xu2019,Morris89,Garg65,Grohe209}, and so their learning and explanatory aspects may differ from general neural networks. Direct learning is expected to be a hard problem in general.

An example of a class of properties going slightly beyond properties considered in~\cite{GNNEx} is having bounded radius~\footnote{The radius of a graph is the smallest number $r$ such that for some vertex $v$ every other vertex can be reached from $v$ by a path of length at most $r$.}. Here we assume that graphs are classified as positive if their radius is at most $r$, for an unknown value of $r$. In this case learning $r$ directly is straightforward, but from the point of view of explainability it is of interest to see if a learned GNN provides meaningful explanations. For graphs classified positive a natural explanation is a spanning tree of bounded depth. Thus the relevant subgraphs do not have bounded size. Nevertheless, the GNNExplainer algorithm can be used with a minor change in the denoising part, which remains computationally efficient.

\begin{figure}
    \centering
    \includegraphics[width=0.355\textwidth]{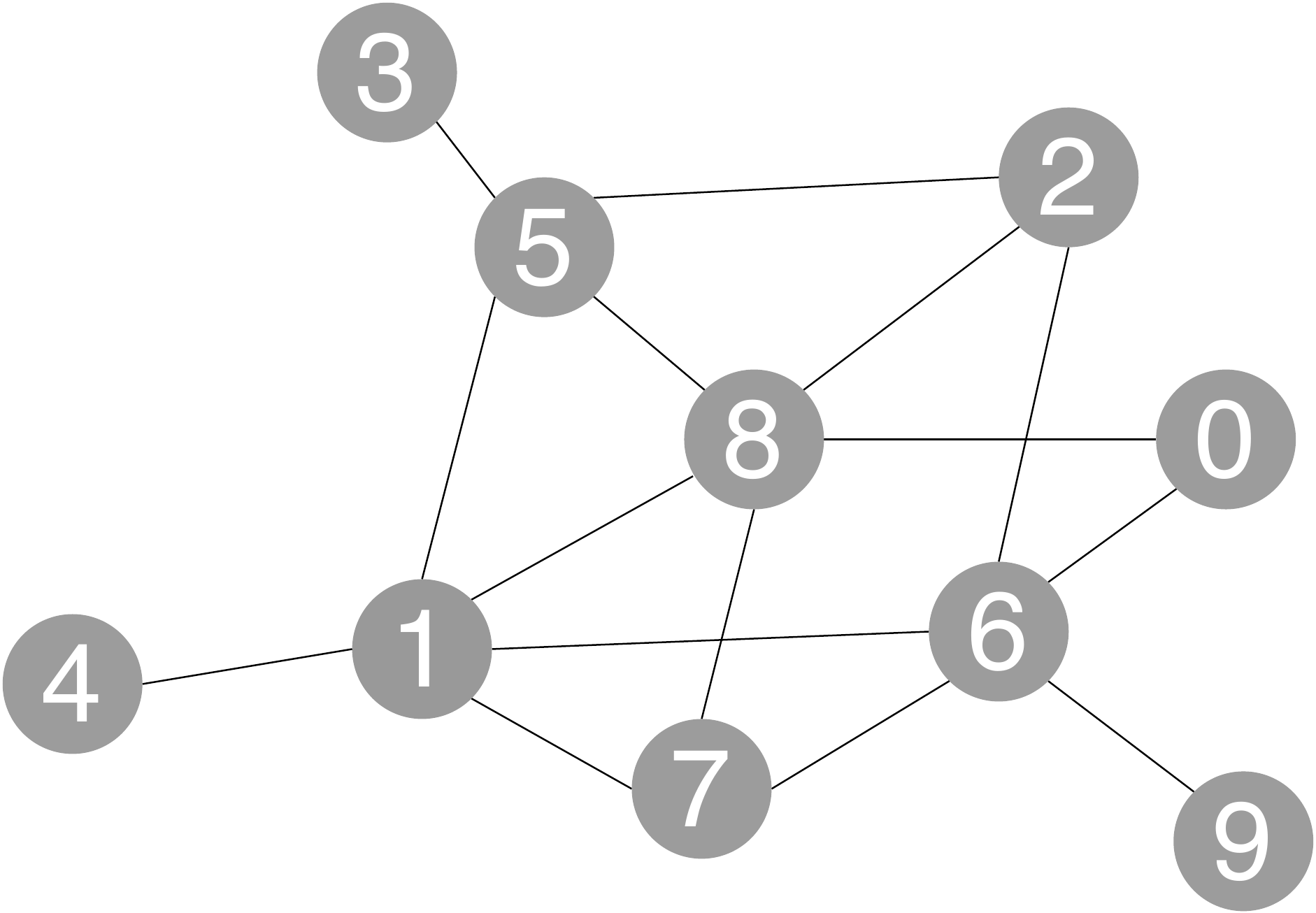}
    \vspace{-0.25cm}
    \caption{\footnotesize{Input graph: \# vertices=10, \# edges=15, radius=2}}
    \label{fig:original_graph}
    \includegraphics[width=0.355\textwidth]{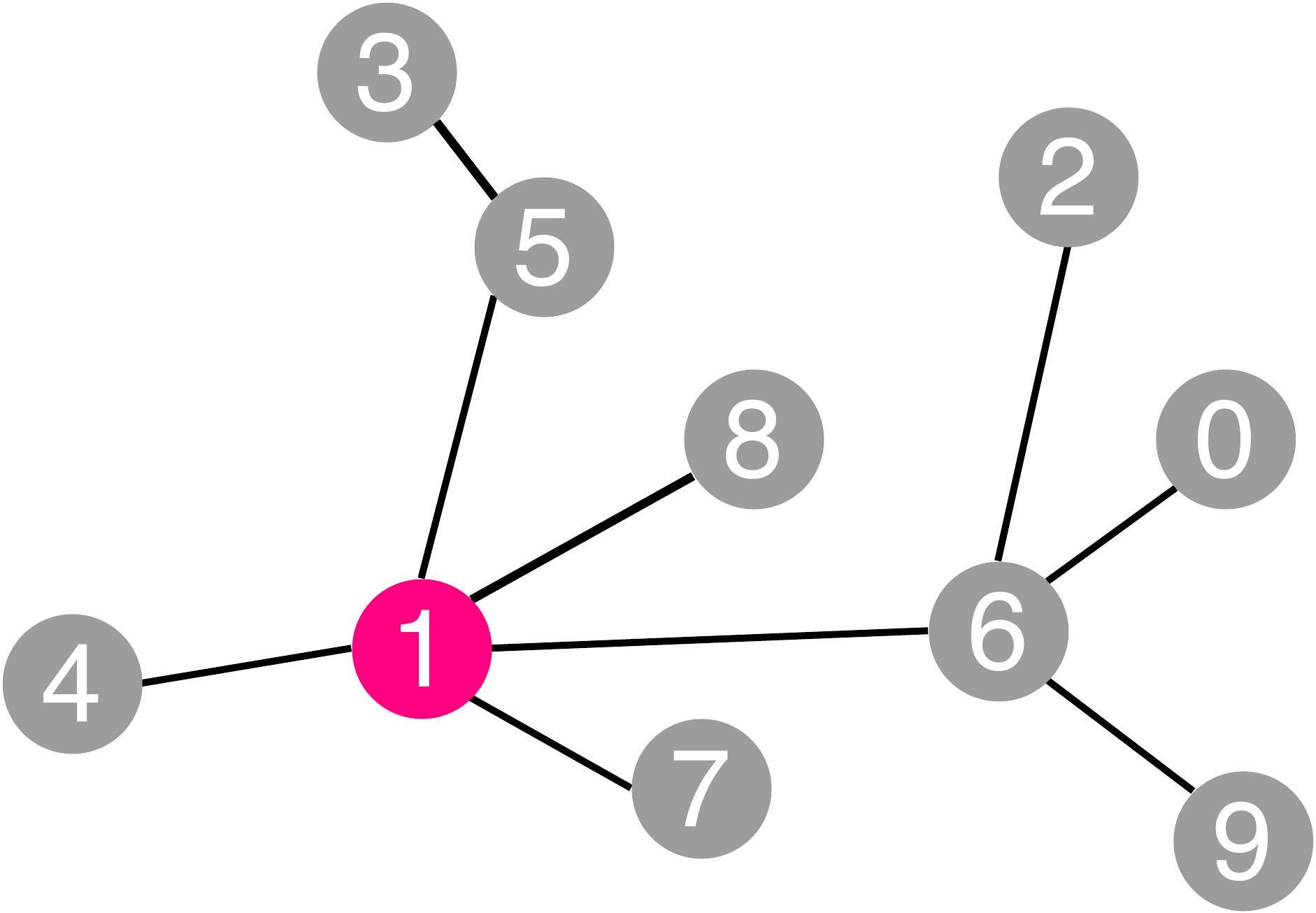}
    \vspace{-0.25cm}
    \caption{\footnotesize{Explanation graph: spanning tree explanation extracted from Figure \ref{fig:original_graph} with vertex $1$ as the center}}
    \label{fig:exp_graph}
\end{figure}

Figures \ref{fig:original_graph} and \ref{fig:exp_graph} give an example of the explanation found in our ongoing experiments for the case $r = 2$.
GNNExplainer assigns weights to the edges reflecting their importance for the classification. The denoising procedure (which differs from the one in~\cite{GNNEx}) then selects an explanatory subgraph. This procedure is designed knowing that the scientist is interested in the role of the radius. There are several options to extract a radius-related explanation from the weighted subgraph, and the preferred one could be chosen through
interaction with the biologist.
The syntactic data assume an idealized case, and it needs to be explored how robust the explanations are if the classification is not in terms of the radius alone.

\section{Prediction without explanation?}

ML providing \emph{prediction without explanation}
in the natural sciences raises fundamental questions in the philosophy of science.
An upcoming special issue of \emph{Minds and Machines} is devoted to these questions~\footnote{https://www.springer.com/journal/11023/updates/18180316 .}. 
Some questions are similar to those about the societal aspects of ML, such as policy decisions and trust, and some are specific to science, such as the effect of changing types and roles of explanations.
The frequent mention in the scientific literature of lack of interpretability of results produced by ML indicates that the problems are present on the level of scientific practice as well.

A major dividing line in the philosophy of science is between realism and anti-realism. Realism,
the dominant approach, asserts that the goal of scientific research is to produce true theories.
Van Fraassen's anti-realist approach claims that the goal is \emph{empirical adequacy} or ``saving the phenomena''.
On the other hand, theories and explanations do have a role in his approach as well. For theories, he emphasizes their semantic nature as a collection of models, as opposed to the syntactic version as a collection of true sentences. His notion of explanations has been discussed above. 

\cite{Hooker2} discuss prediction without explanation versus interpretable modelling in the context of the realism versus anti-realism debate. In their conclusion,
ML leads to ``a mode of scientific advance that is alien to our present philosophical conceptions''. 

Explainability from specification can be seen as an attempt to bridge the gap between ML and the philosophy of science.
It is based on van Fraassen's constructive empiricism, and, in particular, on his pragmatic theory of explanation. In a sense, van Fraassen's philosophy of science seems to anticipate, and be able to accommodate, the new developments involving ML   
through the role it assigns to theories and explanations and its view of the research process.
\emph{If} explainability from specification turns out to be achievable in ML (at least to some extent) then ML
fits van Fraassen's constructive empiricist approach as a tool.

To put it another way, the gap 
would be bridged by incorporating ML into van Fraassen's philosophy of science. 
\cite{Hooker2} mention ML interpretability as a
form of intuitability. It is not a coincidence that
explainability from specification is applicable for areas such as science, where a more rigorous form of explanation can be expected. Intuitability would apply to the types of explanation considered in most of XAI, describing explanations expected in societal contexts. In that sense the gap bridging we suggest is a meeting in the middle.

A neural network predicting a property of molecules accurately contains empirically adequate knowledge and according to van Fraassen's view that is the main objective.
The scientist \emph{can} also be interested in
whether the network provides relevant explanations according to their relevance relation, and this is the form of explainability considered in this paper. As the scientist's relevance relation is presumably adapted to the problem studied (like subgraphs for molecule classification), explainability may have a better chance to occur than a general purpose notion of explanation (for example, a DN explanation).

Thus the main question suggested by this argument is the following: \emph{to what extent can explainability from specification be achieved in ML scientific applications}?

Positive results on explainability give information on the compatibility of scientific results using empirically adequate predictions from ML with the framework of a theory. 
Correspondence between concepts of a theory and concepts computed at a hidden node is a form of interpretability. Using the query languages mentioned above could be helpful in finding such correspondences.
Negative results, i.e., the lack of interpretability could be a source for \emph{predicate invention} (see, e.g.,~\cite{Hocquette}).

A relevant notion for these considerations is the
{\em Rashomon effect}, i.e., the existence of multiple good models for data, introduced by~\cite{breiman2001}. This phenomenon is explored further in ML, in particular for its relation to interpretability in~\cite{Semenova,Dziu}. 
This effect also seems to match van Fraassen's notion of theories through his semantic approach. \cite{Hancox} notes that the Rashomon effect is argued as providing possibilities for interpretability in \cite{Semenova}, while he views it as relating to the lack of robustness, which is a limitation for relevance to reality.

Thus philosophy of science not only provided a motivation for the approach to XAI proposed in this paper but, in the other direction, it seems that an understanding of the possibilities \emph{and} limitations of explainability  by specification could contribute to answering some of the philosophy of science questions posed by ML.

\section{Concluding remarks} \label{sec:dis}

We described an  approach to explainability based on explicit specifications of the kind of explanations which the user deems relevant.
Candidates for areas where this approach could be realized are {\em natural sciences, medicine and forensic science}, where there are basic forms of explanations, such as argumentation, causality and non-causal forms.

Besides the motivations and justifications mentioned earlier, the possibility to prove theoretical results on explanations in well-defined contexts is another reason to consider this approach. Proving such results is an interesting topic in itself for theoretical AI, but the considerations of the previous section give it additional significance.

\cite{Golea3} initiated a study of the complexity of rule extraction, and
\cite{Barcelo8} proved several hardness results. Obtaining knowledge representations with given tractability properties is related to
{\em knowledge compilation} \cite{DarwicheUj}. Here knowledge representation formalisms, such as DNF and OBDD, are compared with respect to their expressivity, operations supported and efficiently decidable properties. Explainability aspects are discussed in~\cite{Darwiche207}.
As learned models contain errors, it is of interest to consider \emph{approximate} knowledge compilation.
\cite{Chubarian8} prove an approximate knowledge compilation result related to interpretability. 
\cite{Macdonald4} consider explanations with an approximate version of prime implicants.

Besides positive results, there are also results on limitations. \cite{Wegener94} showed that, assuming prime factoring has no non-uniform polynomial time algorithms, there is no data structure for Boolean functions which can represent a version of multiplication in polynomial size and allows for efficient implementation of certain operations. Such negative results would be of interest for explainability as well.

Opening a black box and explaining a computational result in a form comprehensible to the user are two hard problems. 
The requirement to solve two interconnected hard problems at the same time contributes to the difficulties of XAI.
The proposed approach to explainability separates
the algorithm design task and the explainability aspect. This could help by disentangling the cognitive and algorithmic aspects of developing explainable learning algorithms.

\cite{RudinNMI} warns against using explanations provided by black boxes for high stakes decisions.
Even though the scope of this warning is not quite clear,
e.g., for self-driving cars,
it seems that the areas suitable for the proposed approach are less of a concern from this point of view.
The role of ML models in these applications is more of an {\em assistant} rather than a decision maker.
Similarly, explainability has to address adversarial aspects, like ``gaming'' and privacy (see, e.g.,~\cite{Milli19}). Another common feature of the areas mentioned is that these issues seem less relevant there as well.

\medskip

{\bf Acknowledgement} We would like to thank G\'abor Berend, Karine Chubarian, Nick Huggett, M\'ark Jelasity, Jie Liang and Viktor Weiszfeiler for useful discussions.

\bibliography{main}
\end{document}